# Classifying Cooking Object's State using a Tuned VGG Convolutional Neural Network

Rahul Paul
Department of Computer Science and Engineering, University of South Florida, Tampa, Florida, USA

*Abstract*— In robotics, knowing the object states and recognizing the desired states are very important. Objects at different states would require different grasping. To achieve different states, different manipulations would be required, as well as different grasping. To analyze the objects at different states, a dataset of cooking objects was created. Cooking consists of various cutting techniques needed for different dishes (e.g. diced, julienne etc.). Identifying each of this state of cooking objects by human can be difficult sometimes too. In this paper, we have analyzed seven different cooking object states by tuning a convolutional neural network (CNN). For this task, images were downloaded and annotated by students and they are divided into training and a completely different test set. By tuning the vgg-16 CNN 77% accuracy was obtained. The work presented in this paper focuses on classification between various object states rather than task recognition or recipe prediction. This framework can be easily adapted in any other object state classification activity.

## I. INTRODUCTION

Food is one of the few basic things that human needs in their daily life. Human's feelings, emotions and health is often correlated with food.

Preparation of a meal to consume by cooking has various sub-stages. And cutting the ingredients into different size, shape or state is one of the preliminary stages of cooking. Ingredients size or state will vary (e.g. diced, juiced, julienne etc.) based on the dishes we are going to make.

For a skilled chef, recognition of those states is easier but to a normal person identifying the state of a cooking object is not an easy task. For example, in Figure 1, two images are shown, one is a grated vegetable and another one is julienne. These two states are difficult to understand and can be confusing sometimes.

In recent years, there have been some work on identifying recipes [10,11] and predicting cooking tasks [7-9] and activity classification [5,6]. But identifying cooking ingredients state detection or classification is a newer task to investigate.

Convolutional Neural Network (CNN) has been used extensively for object classification and recognition recently. Though LeCun's LeNet [1] is one of the first modern neural network architectures, CNN gained popularity in the classification and recognition tasks after the innovation of AlexNet [2] in the ImageNet Competition [3]. In this study, we used vgg-16 CNN architecture [4] for classification.

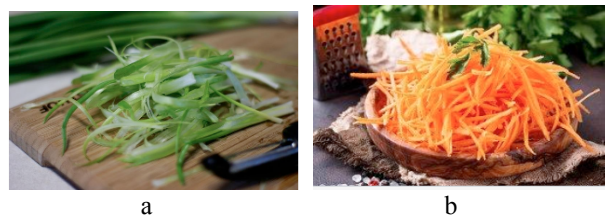

Fig. 1. (a) Example of Julienne, (b) Example of Grated

This paper provides baseline results for cooking object state classification from images. We have divided the cooking object states into seven different categories: Whole, Juiced, Sliced, Diced, Creamy paste, Julienne, and Grated. A classification model that performs reliable in recognizing cooking object states is useful in many ways. Firstly, we can incorporate this model into activity recognition of what's cooking to get better understanding and prediction. Secondly, with the emergence of robotics, the robot needs to detect the object state to learn and then choose a cooking object (e.g. grated or sliced from all the seven different options) that the robot will use to do cooking. And lastly, different objects have different consistency, e.g. the whole food vs the creamy paste. So, the robots need to learn the object at different states to improve its grasp.

The remainder of the paper is organized as follows: section 2 discusses previous work regarding activity classification and predicting recipes and cooking states, section 3 introduces convolutional neural network and architecture used in this study, section 4 presents the dataset and result obtained. Finally, section 5 discuss on present and future work and concludes the paper.

## II. PREVIOUS WORK

In activity classification and prediction of cooking tasks, Lester [5] used discriminative useful features for an ensemble of classifiers with HMMs to recognize ten different actions (e.g. running, jogging, walking etc.) from a multimodal dataset. Spriggs [6] extracted frames from a video and projected the features from a frame to a lower dimensional space using PCA and for clustering GMM was used and HMMs to classify two cooking activity videos. Lade [7] tried to recognize cooking activities using Hierarchical state space Markov model (which models temporal relations between the different tasks) and Object based task grouping (OTG) (it takes account relation between tasks based on usage of common objects). For interpreting food recipes, Bossard [8] introduced Food-101 dataset with a baseline classification of 50.8% accuracy. They used random forests to find discriminative areas simultaneously for all classes. Liu [9] proposed a CNN based for recognition approach on Food 101 dataset and improved the baseline accuracy to 77.4%. Myers [10] presented a system to recognizing the contents of a meal from an image and then predict its calories. They also improved the classification accuracy further on Food -101 dataset to 79% and came up with a mobile phone based application. Malmaud [11] proposed a novel method for interpreting cooking videos using text, speech and vision. HMM was used to make the speech and recipe steps in align and then deep convolutional neural network was used to detect the food.

Our current study was focused on recognizing objects at different states to help the robots recognize and improve its grasp for different object states. In robotics, knowing the object states and recognizing archiving the desired states are very important. Sun [17,18] presented a novel object learning approach for robots to understand the object's interaction from human. Bayesian network was used to represent the knowledge learned and using this network the recognition reliability of objects and human could improve, which would in turn help the robots to properly understand a pair of objects. Paulius [19] presented FOON (functional object-oriented network), a novel knowledge structure representation to analyze the connectivity object's motion in manipulation tasks. FOON was learned using change of object states and manipulations of objects by human. Using the FOON, the robot could understand and decode the goal of a task and identify objects at different states on which to operate and generate a proper sequence of manipulation motion. Robots also need to learn how to grasp or handle objects at different state separately. To achieve different states, different manipulations would be required, as well as different grasping. Lin [20] proposed a fingertip force learning algorithm for grasping with the help of a human teacher. A machine learning based procedure based on gaussian mixture model was applied on fingertip force and find the position to obtain the motion and force model and then by applying gaussian mixture regression a motion trajectory was obtained. In another study by Lin [21,22], demonstrated an approach to utilize position of thumb for different grasp type and orientations on grasped object for a grasp planning procedure. They evaluated their approach on 8 daily objects with a Barrett hand and a Shadow hand. The difficulty of task oriented grasp planning was task modelling. Lin [23] modeled a manipulation task by constructed a statistical model from disturbance data. Huang [24] focused on the requirements of grasps from the physical interactions in instrument manipulations. The requirements of manipulation-oriented grasp were associated directly to the instrument functionality and the manipulated task and not associated with the robotic hardware.

In this paper, we proposed an approach to classify the object state of cooking ingredients, which hasn't been studied yet.

III. CONVOLUTIONAL NEURAL NETWORK

Convolutional neural networks (CNN) are a form of multilayer feedforward neural networks and are emerging for object recognition and classification. In a feed-forward network, the outputs from a previous layer are connected to the next layer. In CNN, convolutional layers are used on top of a feedforward neural network. By using convolutional layers, variations or translations in the data can be handled effectively and more image based features (size, shape, edges etc.) can be obtained by applying different convolutional kernels.

Table. 1. Tuned vgg-16 CNN Architecture-1

| Layers | Weights |
|---|---|
| Input 224x224 RGB image | |
| Conv3-64 | |
| Conv3-64 | |
| Maxpool | |
| Conv3-128 | |
| Conv3-128 | |
| Maxpool | |
| Conv3-256 | |
| Conv3-256 | |
| Conv3-256 | |
| Maxpool | |
| Conv3-512 | |
| Conv3-512 | |
| Conv3-512 | |
| Maxpool | |
| Conv3-512 | Weights from base model was used till this layer |
| Conv3-512 | |
| Conv3-512 | |
| Conv3-512 | Added to the Network |
| Maxpool | |
| Fully Connected 1-4096 Dropout 0.2 Fully Connected 2 -128 Dropout 0.2 Fully Connected 3 - 7 | Changed the FC layers and Dropout was added. |

A CNN can be developed using different layers; convolutional layers (extracting low-level information from images using convolutional kernels), pooling layer (reduce data dimensionality), activation function (adding non-linearity to CNN) and fully connected layers (feed forward network layers) [12]. In this study, we tuned vgg-16 architecture [4] for classification. Vgg-16 architecture was trained on ImageNet [3] dataset. We only tuned the weights of the top layers of

vgg-16 architecture and kept the weights of the bottom layers same using Keras [13] with Tensorflow [14] backend. Weights of the lower layers were kept same as the lower layers will grasp only size, shape, edges, textures etc. Tuned vgg-16 is described in Table 1 and 2. We had only ~5000 original images, so to avoid overfitting dropout [15] was used. RMSprop [16] was used as the gradient descent optimization algorithm. Learning rate was kept constant at 0.0001. Batch size of 16 was used for both training and validation. A total of 50 epochs were used to train the CNN. A few filters from the vgg-16 are shown in Figure 2.

Table. 2. Tuned vgg-16 CNN Architecture-2

| Layers | Weights |
|---|---|
| Input 224x224 RGB image | |
| Conv3-64 | |
| Conv3-64 | |
| Maxpool | |
| Conv3-128 | |
| Conv3-128 | |
| Maxpool | |
| Conv3-256 | |
| Conv3-256 | |
| Conv3-256 | |
| Maxpool | |
| Conv3-512 | |
| Conv3-512 | |
| Conv3-512 | |
| Maxpool | |
| Conv3-512 | Weights from base model was used till this layer |
| Conv3-512 | |
| Conv3-512 | |
| Conv3-512 (3x3 kernel) | Added to the Network |
| Maxpool | |
| Conv3 -512 (1x1 kernel) | |
| Global Average Pooling | Changed the FC layers and Dropout was added. |
| L2 regularization (0.01) | |
| Dropout 0.2 | |
| Fully Connected 1 - 7 | |

## IV. EXPERIMENT AND RESULT

The dataset for our study was generated by downloading different states of cooking images from internet. The dataset (version 1.0) can be accessed using the given link (http://rpal.cse.usf.edu/datasets_cooking_state_recognition.html). These images were further distributed among students in class for labelling and then each student's results were verified by the TA and other students. Our training dataset has 5117 images from 7 different class (Whole, Juiced, Sliced, Diced, Creamy paste, Julienne, and Grated). A fully separated test set of images was used to verify the best trained model [26]. The size of the training dataset was smaller to train, so external image augmentation by rotation (45 degree) and flipping (both horizontally and vertically) was applied. After the image augmentation, 70% of the training images were used to train the CNN, 20% of the remaining images were used to validate the model. And the remaining 10% images were used to test the classification power of the model and based on the result of that test result the parameters of the CNN was changed. RMSprop stochastic gradient descent algorithm was used to search the optimum converging point faster than a normal SGD (stochastic descent algorithm).

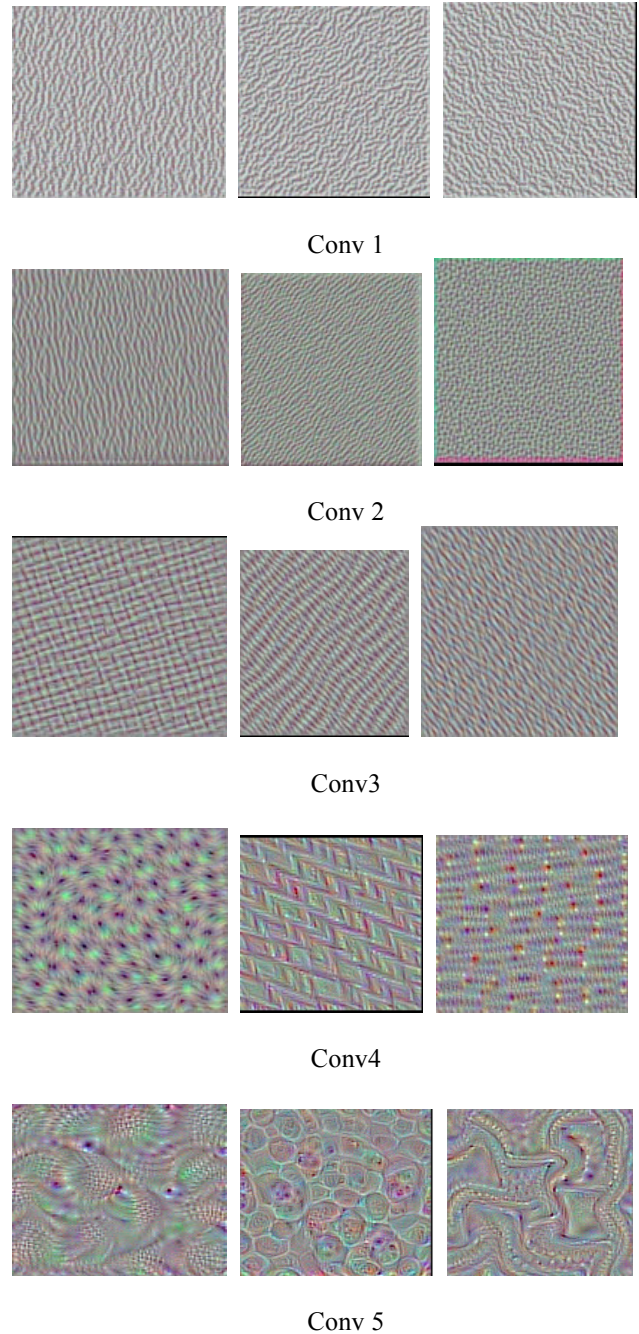

Conv 1

Conv 2

Conv3

Conv4

Conv 5

Fig. 2. Example of few filters from vgg-16 CNN

The size of the input images for vgg-16 was 224x224, so bi-linear interpolation was applied for resizing the images.

The best performing model was used to check the unseen test set and obtained 77% accuracy. From the first tuned CNN (CNN architecture-1) model 77% accuracy was obtained, whereas from the second tuned model (CNN architecture-2) 76.6% accuracy was achieved over an unseen test dataset. Loss function and accuracy for training and validation for our best model (CNN architecture-1) are shown in Figure 3. Some of the misclassified images are shown in Figure 4.

to stop over-fitting. From the loss plot, we found that though our trained model was not over-fitted in the beginning (till 10 epochs) but after that it became little over-fitted. To reduce overfitting data augmentation was applied. But there was a huge difference between the total augmented images and the total weights of the tuned CNN architecture. Because of that, the tuned architecture became over-fit. To reduce the overfitting and training weights of the CNN, another tuned CNN architecture was proposed.

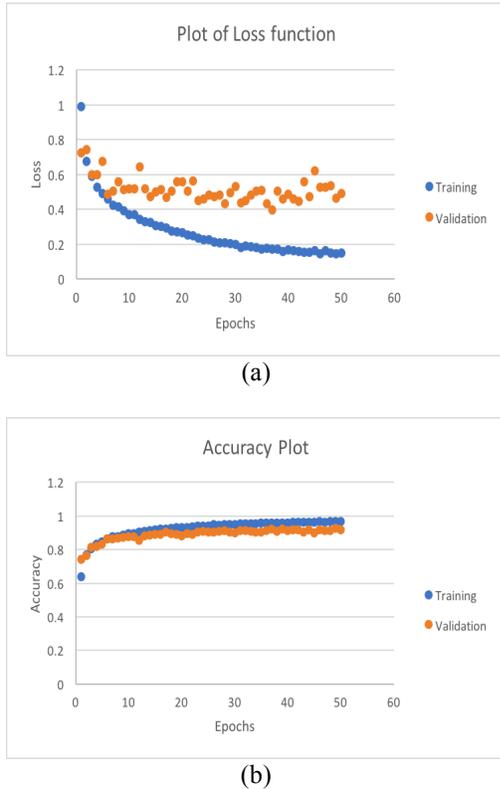

(a)

(b)

Fig. 3. Loss and Accuracy plot of training and validation

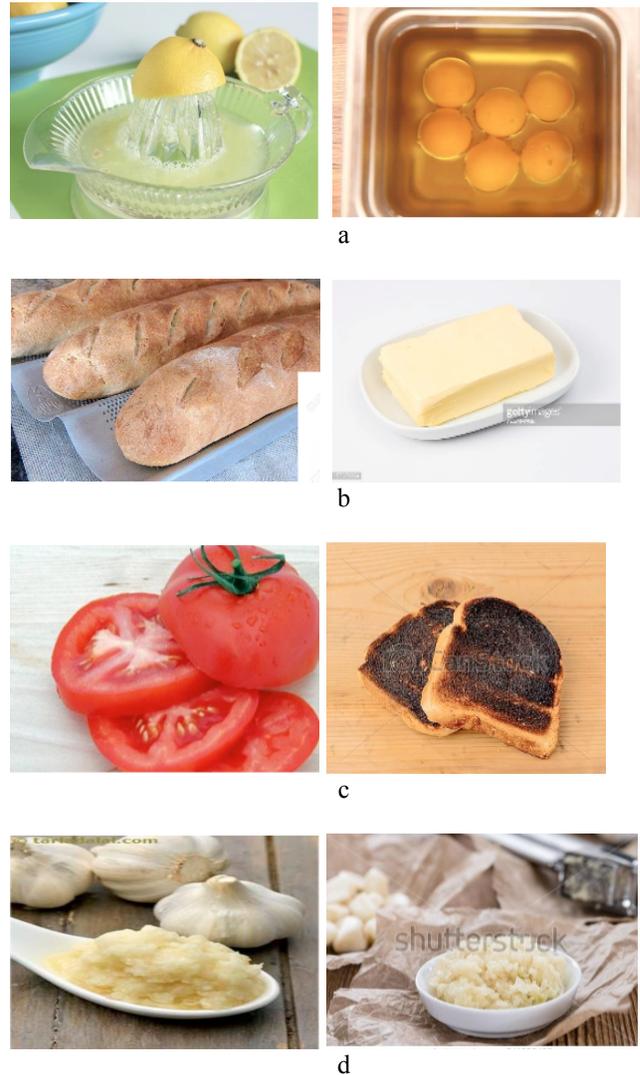

Fig. 4. Misclassified examples: (a) Original label is juiced, classified label is sliced, (b) original label is whole, classified label is sliced, (c) original label is sliced, classified as whole, (d) original label was grated, classified as creamy

## V. DISCUSSIONS

Knowing the object states and recognizing the desired states are very important. objects at different states would require different grasping. To achieve different states, different manipulations would be required, as well as different grasping. For achieving this task, a cooking object dataset consisted of seven different states were chosen. Pre-trained vgg-16 convolutional neural network was utilized to tune the parameters and re-train using our dataset. In this paper, we experimented with two tuned architectures, though CNN architecture 1 slightly performed better than CNN architecture 2. In the first CNN architecture, we used all the bottom layers till the fully connected layers. Another convolution layer was added before applying the max-pooling layer. The fully connected layer was modified too. First fully connected layer had 4096 neurons, but the second fully connected layer had 128 neurons before the final classification layer of 7 neurons. Dropouts were applied in between the fully connected layers

In this architecture also, another convolutional layer was added before the max-pooling layer. In place of fully connected layer, we applied 1x1 convolution. 1x1 convolution was used to reduce dimensionality and more non-linearity was added to the feature representation learned by the previous layers to enhance network representations. Global Average pooling layer was used after the 1x1 convolution layer to reduce the total number of parameters of the model and spatial dimension. To reduce overfitting dropout along with L2 regularization was applied. Srivastava [25] showed that L2 regularization along with dropout enhanced the model prediction performance and reduce overfitting. We found that CNN architecture-2 had less over-fit than the CNN architecture-1, though the best performed model's loss was shown in the paper.

## VI. CONCLUSIONS

In this study, we proposed a baseline result for state of object classification using CNN on cooking objects of different states. The cooking objects has seven different states: whole, sliced, juiced, grated, julienne, creamy paste and diced. Tuned Vgg-16 model was utilized for this study. The training dataset had 5117 images and a separate unseen test dataset was used to obtain the classification performance. 77% accuracy was obtained using the best trained model on unseen test data. This study was challenging because there were some images (figure 1) where the object state can create a confusion for normal people by looking at an image. Previously, there were several studies on cooking activity recognition and identifying recipes. But object state classification was a new topic of research. Object state classification can be utilized in future in robotics. E.g. the robot needs to detect the object state to learn and then choose a cooking object (e.g. grated or sliced from all the seven different options) that the robot will use for cooking. In future study, a larger dataset is needed for better analysis and other newer CNN architectures will be utilized for classification. Ensemble of CNNs will be evaluated as well.